%% file: main.tex
\documentclass[10pt,twocolumn,letterpaper]{article}

\usepackage{iccv}
\usepackage{times}
\usepackage{epsfig}
\usepackage{graphicx}
\usepackage{amsmath}
\usepackage{amssymb}
\usepackage{multirow}
\usepackage[dvipsnames]{xcolor}
\usepackage[normalem]{ulem}
\usepackage{soul}
\usepackage{enumitem}
\usepackage{bbm}
\usepackage{subcaption}
\usepackage{pifont}
\usepackage{booktabs}
\usepackage{tabularx}
\usepackage[export]{adjustbox}
\usepackage{caption,stackengine}
\usepackage{lipsum}

\usepackage[pagebackref=true,breaklinks=true,letterpaper=true,colorlinks,bookmarks=false]{hyperref}

\usepackage[capitalize]{cleveref}
\crefname{section}{Sec.}{Secs.}
\Crefname{section}{Section}{Sections}
\Crefname{table}{Table}{Tables}
\crefname{table}{Tab.}{Tabs.}
\iccvfinalcopy 

\def\iccvPaperID{5933} 

\newcommand{\lyc}[1]{\textcolor{black}{#1}}

\newcommand{\red}[1]{{\color{red}{#1}}}
\newcommand{\green}[1]{{\color{OliveGreen}{#1}}}
\ificcvfinal\fi

\input{preamble}

\begin{document}

\title{Discovering Spatio-Temporal Rationales for Video Question Answering}

\author{Yicong Li$^1$, Junbin Xiao$^1$, Chun Feng$^2$, Xiang Wang$^2$, Tat-Seng Chua$^1$\\
$^1$National University of Singapore, 
$^2$University of Science and Technology of China,\\
{\tt\small liyicong@u.nus.edu,fengchun3364@mail.ustc.edu.cn, xiangwang1223@gmail.com} \\
{\tt\small junbin@comp.nus.edu.sg,  dcscts@nus.edu.sg}
}

\maketitle
\ificcvfinal\thispagestyle{empty}\fi

\input{sec/0_abstract}

\input{sec/1_intro}

\input{sec/related}

\input{sec/2_preliminary}

\input{sec/3_method}
\input{sec/4_exp}

\input{sec/5_conclusion}

{\small
\bibliographystyle{ieee_fullname}
\bibliography{main}
}

\end{document}

%% file: preamble.tex
\usepackage{overpic}
\usepackage{enumitem} 
\usepackage{overpic} 
\usepackage{color}

\definecolor{turquoise}{cmyk}{0.65,0,0.1,0.3}
\definecolor{purple}{rgb}{0.65,0,0.65}
\definecolor{dark_green}{rgb}{0, 0.5, 0}
\definecolor{orange}{rgb}{0.8, 0.6, 0.2}
\definecolor{red}{rgb}{0.8, 0.2, 0.2}
\definecolor{darkred}{rgb}{0.6, 0.1, 0.05}
\definecolor{blueish}{rgb}{0.0, 0.3, .6}
\definecolor{light_gray}{rgb}{0.7, 0.7, .7}
\definecolor{pink}{rgb}{1, 0, 1}
\definecolor{greyblue}{rgb}{0.25, 0.25, 1}

\usepackage{blindtext}

\renewcommand{\paragraph}[1]{\vspace{1em}\noindent\textbf{#1}.}

\newcommand{\Mat}[1]{\textbf{#1}}

\newcommand{\Space}[1]{\mathbb{#1}}

%% file: sec/0_abstract.tex
\begin{abstract}
This paper strives to solve complex video question answering (VideoQA) which features long video containing multiple objects and events at different time. To tackle the challenge, we highlight the importance of identifying question-critical temporal moments and spatial objects from the vast amount of video content. Towards this, we propose a \textbf{S}patio-\textbf{T}emporal \textbf{R}ationalization (STR), a differentiable selection module that adaptively collects question-critical moments and objects using cross-modal interaction.
The discovered video moments and objects are then served as grounded rationales to support answer reasoning. Based on STR, we further propose TranSTR, a Transformer-style neural network architecture that takes STR as the core and additionally underscores a novel answer interaction mechanism to coordinate STR for answer decoding. Experiments on four datasets show that TranSTR achieves new state-of-the-art (SoTA). Especially, on NExT-QA and Causal-VidQA which feature complex VideoQA, it significantly surpasses the previous SoTA by 5.8\% and 6.8\%, respectively. We then conduct extensive studies to verify the importance of STR as well as the proposed answer interaction mechanism. With the success of TranSTR and our comprehensive analysis, we hope this work can spark more future efforts in complex VideoQA.  
Code will be released at \url{https://github.com/yl3800/TranSTR}.

\vspace{-0.5cm}
\end{abstract}

%% file: sec/1_intro.tex
\section{Introduction}\label{sec:intro}
The great success of self-supervised pretraining with powerful transformer-style architectures \cite{devlin2018bert,deberta,DBLP:conf/icml/KumarIOIBGZPS16,fu2021violet,yang2021just,zellers2021merlot} has significantly boosted the performance of answering simple questions (\eg, ``what is the man doing'') on short videos (\eg, 3$\sim$15s) \cite{jang2017tgif,DBLP:conf/mm/XuZX0Z0Z17}. The advances thus point towards complex video question answering (VideoQA), \lyc{that features long video containing multiple objects and events} \cite{next-qa,causalvid,zhong2022video}. Compared with simple VideoQA, complex VideoQA poses several unique challenges:

\begin{figure}
  \centering
  \begin{subfigure}{\linewidth}
  \includegraphics[width=1\linewidth]{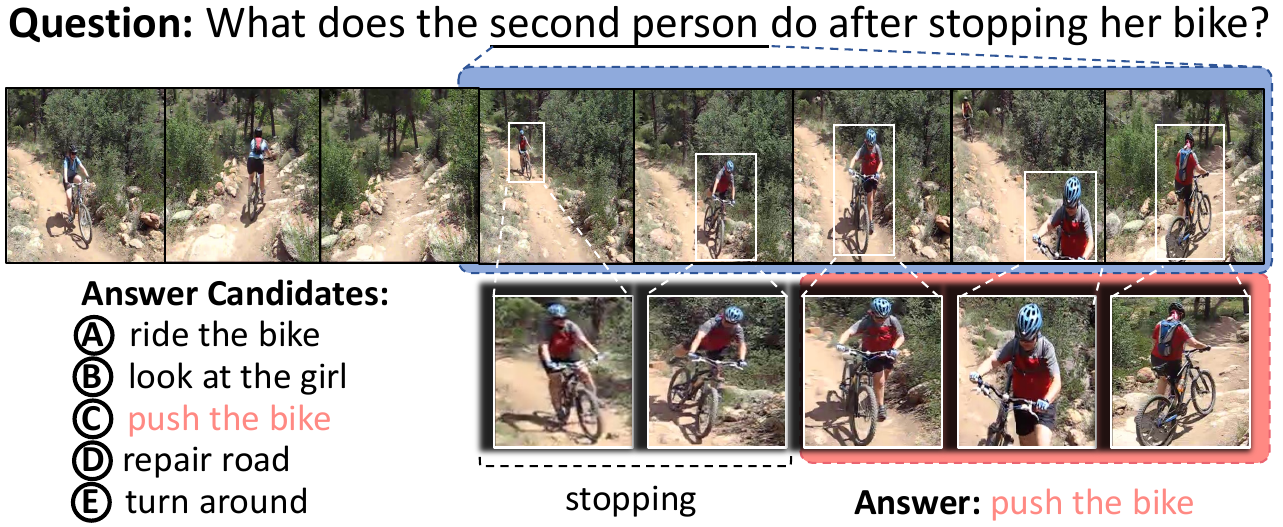}\par
  \caption{A example of long video (52s) with multiple objects, the question-related frames and objects are located to support the reasoning.}
  \label{fig:1a} 
  \end{subfigure}

  \begin{subfigure}{0.495\linewidth}
  \includegraphics[width=\linewidth]{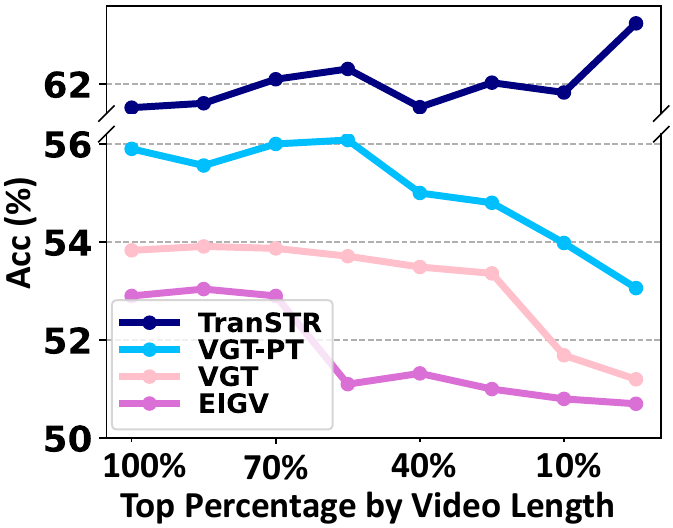}
  \caption{Accuracy by video length.}
  \label{fig:1b} 
\end{subfigure}
\begin{subfigure}{0.485\linewidth}
  \includegraphics[width=\linewidth]{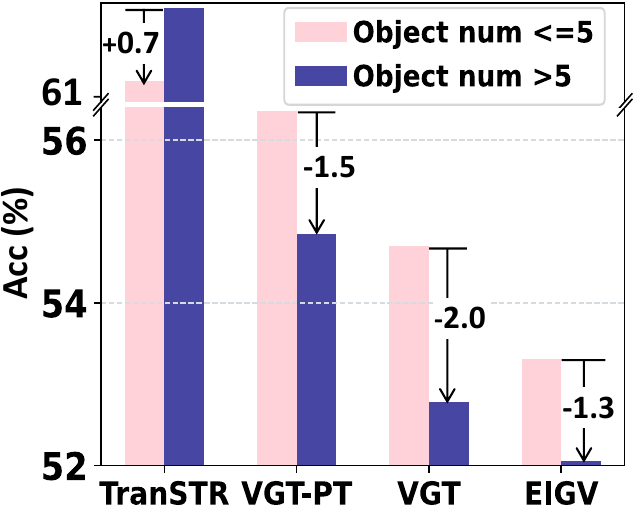}
  \caption{Accuracy by object number.}
  \label{fig:1c}
\end{subfigure}
\caption{(a) Illustration of complex VideoQA, in which the videos are longer and the questions involve multiple objects and events at different time. (b) \lyc{Prediction accuracy grouped by video length. We first sort all samples by video length, then select top x\% to calculate accuracy.}
(c) Accuracy grouped by whether the video has more than 5 objects. All results are reported on NExT-QA test set \cite{next-qa}. Figure (b) and (c) show that our method TranSTR performs much better than the previous SoTAs for question answering of long videos with multiple objects.}
\label{fig:1}
\vspace{-0.6cm}
\end{figure}

1) \textbf{Longer videos with multiple objects interacting differently at different time.} The long video and rich visual content indispensably bring more background scenes that include  massive question-irrelevant video moments and objects. For example, to answer the question in \cref{fig:1a}, only the interaction between object ``person'' and ``bike'' on the last three frames encloses the answer information, leaving the massive rest as background. These backgrounds, if not filtered properly, will overwhelm the critical scene and interfere with answer prediction. 
%
2) \textbf{Harder negative answer as distractors.}
Negative answers in complex VideoQA are typically tailored for each video instance. Due to the massive video content, the vast question-irrelevant scene provides an ideal foundation to build a hard negative candidate as distractor.
The hard negatives are very similar to the correct answer but correspond to a different video moment or object.
For example, the answer candidate ``A.ride the bike'' of \cref{fig:1a}, though irrelevant to the question,  corresponds to a large part of the video. 
As a result, these distractors can seriously derail the prediction if not properly modeled.

In light of the challenges, current methods (both pretrained and task-specific architectures) hardly perform well on complex VideoQA. 
In \cref{fig:1b,fig:1c}, we use the video length and number of objects\footnote{We acquire the object number using annotation of video relation detection dataset \cite{vidvrd}, which shares same source video as NExT-QA.} to indicate the complexity of the video questions. We can see that current methods suffer a drastic performance drop when video length increases or more objects are involved. The reason can be provided from two aspects:
%
%
\textbf{First}, confronting long video and multiple objects, pretrained methods suffer from a domain gap. Because they are typically pretrained with short videos and simple questions, \cite{DBLP:conf/icml/KumarIOIBGZPS16, lei2021less} 
where the answer can be easily captured via a static frame, without fine-grained reasoning over multiple objects in a long video. 
While recent task-specific methods exploit object-level representation for fine-grained reasoning \cite{hostr,pgat,hqga,VGT}, they exhibit limited generalization ability, as they handle different videos with only a fixed number of frames and objects, and cannot adapt to lengthy and varied visual content,
which rigidity undermines their adaptability to a wide range of video content.
\textbf{Second}, to model the answer candidate, prevailing designs \cite{jang2017tgif,gao2018motionappearance,fan2019heterogeneous,hga,hqga} append the candiate to the question and treat the formed question-answer sequence as a whole for cross-modal learning. However, this makes the answer candidate directly interact with the whole video content, which gives rise to a strong spurious correlation between the hard negative candidates (\eg ``A. ride the bike" in \cref{fig:1a}) and the question-irrelevant scenes (\eg `` riding'' scene in first three frames), leading to a false positive prediction. 
%

In this regard, we propose TranSTR, a Transformer-style VideoQA architecture that coordinates a Spatio-Temporal Rationalization (STR) with a more reasonable video-text interaction pipeline for candidate answer modeling.
%
%
%
STR first temporally collects the critical frames from a long video, followed by a spatial selection of objects on the identified frames. By further fusing the selected visual objects and frames via light-weight reasoning module, we derive spatial and temporal rationales that exclusively support answering. 
%
In addition to STR, we circumvent the spurious correlation in the current modeling of answer candidates by formulating a more reasonable video-text interaction pipeline, where the question and answer candidates are separately (instead of being appended as a whole) fed to the model at different stages. 
Specifically, before rationale selection, only the question is interacted with the video to filter out the massive background content while keeping question-critical frames and objects. After that, a transformer-style answer decoder introduces the answer candidates to these critical elements to determine the correct answer.
Such a strategy prevents the interaction between the hard negative answers and the massive background scenes, thus enabling our STR to perform better on complex VideoQA (see TranSTR in \cref{fig:1b,fig:1c}). 
\lyc{It is worth noting that STR and the answering modeling are reciprocal. Without STR's selection, all visual content will still be exposed to answer candidates. Without our answer modeling, STR could identify the question-irrelevant frame and object as critical.
Thus, the success of TranSTR is attributed to the integration of both.}

Our contributions are summarized as follows:

\begin{itemize}[leftmargin=*]
    \item We analyze the necessity and challenge of complex VideoQA. To solve the task, we identify the importance of discovering spatio-temporal rationales and preventing spurious correlation in modeling candidate answers.
    \item We propose TranSTR that features a spatio-temporal rationalization (STR) module together with a more reasonable candidate answer modeling strategy. The answer modeling strategy is independently verified to be effective in boosting other existing VideoQA models.
    \item We perform extensive experiments and achieve SoTA performance on four popular benchmarks, especially for ones that features complex VideoQA (NExT-QA \cite{next-qa} +5.8\%, CausalVid-QA \cite{causalvid} +6.8\%)
\end{itemize}

%% file: sec/related.tex
\section{Related Works}\label{sec:related}

\noindent\textbf{Video Question Answering (VideoQA).}
Substantiated as a fundamental extension of ImageQA, VideoQA has enlarged its definition by adding a temporal extension. 
According to the pre-extracted feature granularity, existing methods either use the frame-level features or incorporate object features for fine-grain reasoning. 
In task-specific designs, Focus on simple questions and short videos, earlier efforts tend to model the video sequence as a visual graph using purely frame features. As the pioneer of graph-based structure, \cite{hga} and \cite{park2021bridge} build their typologies based on the heterogeneity of input modality,  while \cite{mspan} enables progressive relational reasoning between multi-scale graphs. 
Recently, the emergence of complex VideoQA benchmarks \cite{next-qa, causalvid} has prompted studies on long video with multiple visual entities. 
In this regard, Another line of research has prevailed by processing video as multi-level hierarchy. 
\cite{hcrn} first build a bottom-up pathway by assembling information first from frame-level, then merging to clip-level. The following works \cite{hostr, hqga} extend the hierarchy into the object-level, where a modular network is designed to connect objects on the same frame. Most recently, \cite{VGT} establish its improvement by enabling relation reasoning in a sense of object dynamics via temporal graph transformer.
Despite effectiveness, the current designs unanimously rely on a fixed number of frames and objects, which severely compromise their transferability across diverse video instances.
In sharp contrast, our method works in a fully adaptive manner to explicitly select frames and objects for reasoning over different circumstances, which demonstrates superior generalization ability.

\vspace{5pt}
\noindent\textbf{Rationalization.}
In pursuit of  explainability, the recent development of DNN is encouraged to reveal the intuitive evidence of their prediction, \ie the rationales. 
As one of the prevailing practices, the rationalization has been extended from the NLP community\cite{DBLP:conf/kdd/Ribeiro0G16} to the Graph \cite{DIR} and Vision field \cite{DBLP:conf/cvpr/ZhangYMW19}. 
Recently, this development also stems from the multi-modal community.
\cite{DBLP:conf/cvpr/ParkHARSDR18} and \cite{DBLP:conf/cvpr/DuaKB21} proposes ImageQA-based tasks that inquire about additional textual evidence, \cite{causalvid} brings this idea to the videoQA.
Despite the progress, the recent solution focus on the rationale only at frame level, and they either require a rationale finder with heavy computation overhead \cite{IGV} or needs to be trained in a data-hungry contrastive manner \cite{EIGV}. TranSTR, however, identifies both critical frames and objects from an efficient cross-modal view. 
Also, distinct from the token reduction method in transformer literature, which trades accuracy for efficiency. Rationalization intends to improve performance \cite{rationalization-robustness}. The intuition behind is that, if a model can find the causal part, they have the potential to ignore the noise. 


%% file: sec/2_preliminary.tex
\section{Preliminaries}
\label{sec:preliminaries}

\vspace{5pt}
\noindent \textbf{Modeling.}
Given the video $V$ and the question $Q$, the VideoQA model $\phi(V,Q)$ aims to encapsulate the visual content and linguistic semantics and choose the predictive answer $\hat{A}$ from the answer candidates.
%
Typically, an entropy-based risk function $\mathcal{L}(\phi(V,Q), A)$ is applied to approach the ground-truth answer ${A}$.

\vspace{5pt}
\noindent \textbf{Data representation.}
We uniformly sample $T$ clips and keep the middle frame of each clip to represent a video.
Then, for each frame, we extract a frame feature $\Mat{f}_t$ via a pretrained image recognition backbone and $S$ object features $\Mat{o}_{t,s}$ using pretrained object detector, where $t,s$ denotes the $s$-th object on the $t$-th frame.
To represent the text, we encode the question as a sequence of $L$ tokens using a pretrained language model and obtain a textual representation $\Mat{q}_l$ for each of them. The visual backbones are frozen during training while the language backbone is finetuned end-to-end as in \cite{VGT}.
To project the representations into a common $d$-dimensional space, we apply a three linear mappings on $\Mat{f}_t$, $\Mat{o}_{t,s}$, and $\Mat{q}_l$, respectively, and thus acquire 
$\Mat{F}\!=\!\left\{ \Mat{f}_t \right\}_{t=1}^{T} \!\in\!\Space{R}^{T\times d}$, 
$\Mat{O}\!=\!\left\{ \Mat{o}_{t,s}\right\}_{t=1,s=1}^{T,S}\!\in\!\Space{R}^{T\times S \times d}$, 
and 
$\Mat{Q}\!=\!\left\{ \Mat{q}_l \right\}_{l=1}^{L}\!\in\!\Space{R}^{L\times d}$ to denote the frame, object, and question features, respectively.

%% file: sec/3_method.tex
\section{Method}\label{sec:method}


\begin{figure*}
\begin{minipage}{0.7\textwidth}
\begin{subfigure}{\textwidth}
\includegraphics[width=\linewidth]{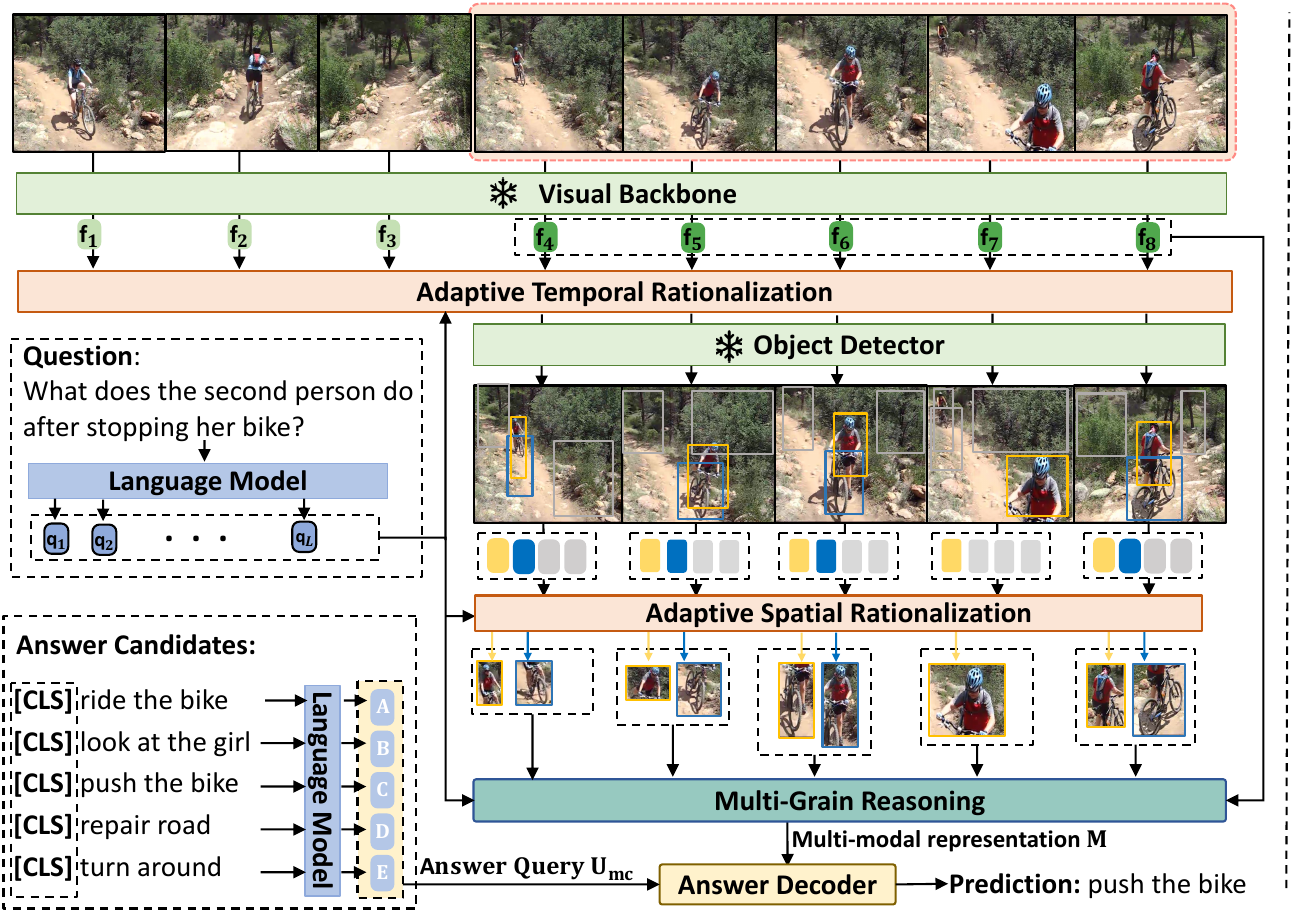}
\vspace{-10pt}
\caption{Overview of TranSTR} \label{fig:2a}
\end{subfigure}
\end{minipage}
\begin{minipage}{0.3\textwidth}
\begin{subfigure}{\textwidth}
\includegraphics[width=1\linewidth]{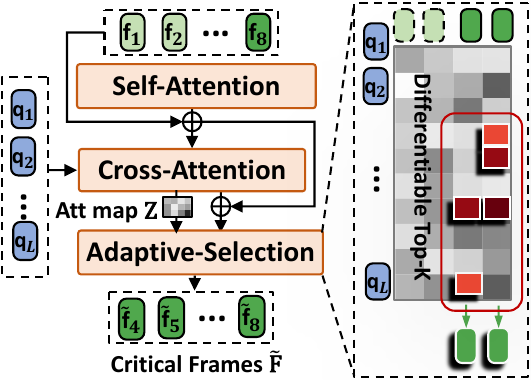}
\vspace{-10pt}
\caption{Adaptive Temporal Rationalization} \label{fig:2b}
\end{subfigure}

\begin{subfigure}{\linewidth}
\includegraphics[width=1\linewidth,]{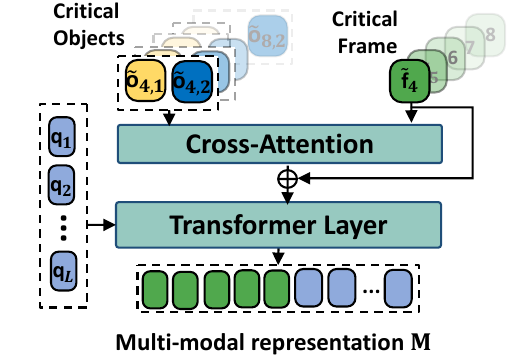}
\vspace{-5pt}
\caption{Multi-Grain Reasoning} \label{fig:2c}
\end{subfigure}
\end{minipage}

\caption{TranSTR (a) contains three components: the Spatio-Temporal Rationalization which adaptively selects critical frames and objects, the Multi-Grain Reasoning (c) which forms multi-modal representation by integrating the critical frames and objects together with the question semantics, and the Answer Decoder which bring in answer candidates and make a prediction based on the multi-modal representation. 
%
Notably, STR follows a ``Temporal then Spatial'' rationalization process, where the two steps leverage a similar deisgn.  In (b), we illustrate this design using Adaptive Temporal Rationalization.}
\end{figure*}

As shown in \cref{fig:2a} TranSTR consists of three main components: Spatio-Temporal Rationalization (STR), Multi-Grain Reasoning (MGR), and Answer Decoder. 
First, STR follows a two-step selection process, where Adaptive Temporal Rationalization (TR) is first  performed for frame selection, followed by object selection via Adaptive Spatial Rationalization (SR).
Next, the selected frames and objects are then combined through MGR, which generates multi-modal representations by fusing with the question.  
Finally, based on the multi-modal representations, the answer decoder takes the combination of answer candidates as queries and predicts an answer.
In this section, we provide a detailed illustration of each module. 

\subsection{Spatio-Temporal Rationalization (STR)}

STR aims to find the question-critical frames and objects in a fully adaptive and differentiable manner, which comprises two components: an Adaptive Temporal-Rationalization (TR) that identifies the critical frames and an Adaptive Spatial-Rationalization (SR) that pinpoints the critical objects on the identified frames.

\subsubsection{Adaptive Temporal Rationalization (TR)}
To identify the critical frames from the video, TR takes the encoded video frames $\Mat{F}\!\in\!\space{R}^{T\times d}$ as input and adaptively selects question-critical frames from a cross-modal view.
As shown in \cref{fig:2b}, a self-attention layer is first adopted to contextualize $\Mat{F}$. Then, a cross-attention is applied by taking contextualized frame feature $\Mat{F}'$ as query and question embedding $\Mat{Q}$ as key and value, which yields the frame tokens $\Mat{F}''\!\in\!\space{R}^{T\times d}$ and the cross-attention map $\Mat{Z}\!\in\!\space{R}^{T\times L}$.
\begin{gather}
    \Mat{F}'=\text{Self-Attention}(\Mat{F}) + \Mat{F}, \label{eq:1}\\
    \Mat{F}'', \Mat{Z}=\text{Cross-Attention}(\Mat{F}', \Mat{Q}) + \Mat{F}'. \label{eq:2}
\end{gather}
For brevity, we omit the superscript in $\Mat{F}''$ and use $\Mat{F}$ to denote the resulting frame tokens. 

Naturally, each value in the cross-attention map indicates the cross-modal interaction activeness between a frame and a question token.
To enable an adaptive frame selection that caters to different videos, we collect the $K_f$ interactions of the highest attention score from the cross-attention map $\Mat{Z}$, then gather their corresponding frame tokens $\tilde{\Mat{F}} \!\in\!\space{R}^{C\times d}$ as a subset of $\Mat{F}$, where $C$ is the number of critical frames. This process is formally given by:
\begin{gather} \label{eq:3}
    \tilde{\Mat{F}}=\text{Adaptive-Selection}_K(\Mat{F}, \Mat{Z}) \quad \text{s.t.}\,K=K_f.
\end{gather}
Notably, by gathering 1-D tokens from the 2-D interaction view, we enable an adaptive collection of $\tilde{\Mat{F}}$ with much fewer tokens being selected as critical frames (\ie $C \!\ll\! T$ and $C \!\ll\! K_f$).
It is worth noting that the interaction selection via vanilla hard Top-K produces a discrete selection, making it inapplicable for end-to-end training. 
We address this issue by adopting a differentiable Top-K using the perturbed maximum method \cite{pertub}, 
which has empirically shown advantages over other differentiable technique (\cf \cref{tab:ablation-loss})

\vspace{-0.6cm}
\subsubsection{Adaptive Spatial Rationalization (SR)}
Given the embedding of the selected frames $\tilde{\Mat{F}}$, SR aims to pinpoint the question-critical objects in each frame. 
To achieve that, SR enable an adaptive object selection similar to \cref{eq:1,eq:2,eq:3}.
Specifically, for the $t$-th critical frame $\tilde{\Mat{f}}_t$, we first feed SR with fixed $S$ object features detected on that frame, then collect top $K_o$ interactions from its cross-modal attention map with question embedding. Finally, by gathering their corresponding object tokens, we obtain the critical object feature $\tilde{\Mat{o}_t} \!\in\!\space{R}^{C_t\times d}$, where $C_t$ denotes the number of critical objects on $t$-th critical frame.
%
It is worth noting that, SR is applied independently to each frame, thus, different frames can adapt to different numbers of the critical objects $C_t$, even if we keep $K_o$ constant for all frames.

\subsection{Multi-Granularity Reasoning (MGR)}
MGR aims to enhance the frame-level representation with fine-grained object embedding, while modeling the video dynamic together with question semantics.
%
%
%
As shown in \cref{fig:2c}, MGR first applies intra-frame aggregation via a cross-attention, which takes the frame feature of the $t$-th critical frame $\tilde{\Mat{f}}_{t} \!\ \in\! \space{R}^{1 \times d}$ as query, and all critical objects in $t$-th frame $\tilde{\Mat{o}}_{t} \! \in\! \space{R}^{{C_{t}} \times d} $ as key and value to generate an object-enhanced representation $\mathring{\Mat{f}}_{t}\! \in \! \space{R}^{1 \times d}$ for the $t$-th frame:
\begin{gather} \label{eq:mga}
    \mathring{\Mat{f}}_{t}=\text{Cross-Attention}(\tilde{\Mat{f}}_{t}, \tilde{\Mat{o}}_{t}) + \tilde{\Mat{f}}_{t}.
\end{gather}
By doing so to all $C$ critical frames, we acquire $\mathring{\Mat{F}} \! \in \space{R}^{C \times d}$ as the object-enhanced frame representation for all critical frames.
%
Next, a transformer layer is adopted to establish cross-frame dynamics, which takes in the concatenation of $\mathring{\Mat{F}}$ and question tokens $\Mat{Q}$, and yields multi-modal representations $\Mat{M} \!\in\!\space{R}^{(C+L) \times d}$ for answer decoding:
\begin{gather}
    \Mat{M}=\text{Transformer-Layer}([\mathring{\Mat{F}};\Mat{Q}]),
\end{gather}
where $[;]$ denotes concatenation operation.

\subsection{Answer Decoding}
Existing methods \cite{VGT,IGV,EIGV} concatenate a question with answer candidates. As analyzed in Sec.~\ref{sec:intro}, these methods suffer from a spurious correlation between negative candidates and question-irrelevant video scenes. To circumvent this issue \lyc{in spatio-temporal rationalization}, we employ a transformer-style decoder that takes as input the question-critical multi-modal representations $\Mat{M}$ and the representations of the candidate answers to determine the correct answer. We detail our implementations for multi-choice QA and open-ended QA in the next sub-sections. 

\vspace{-7pt}
\subsubsection{Multi-Choice QA} 
In Multi-Choice QA, answer candidates are given as $\left| A_{mc} \right|$ sentences or short phrases that are tailored for each video-question pair. Therefore, reasoning on Multi-Choice QA typically requires fine-grain inspection of the video content as well as the interaction between the video and the candidate answers.
To this end, we first prepend a $\left[ \text{CLS} \right]$ token to each answer candidate and feed the sequences to the same language model used for question encoding. Then, we gather the output of the `[CLS]' tokens for all encoded answer candidates, and form the answer query $\Mat{U}_{mc} \in \Space{R}^{\left| A_{mc} \right| \times d}$.
During decoding, we feed a transformer decoder with $\Mat{U}_{mc}$ as query to interact with the multi-modal representation $\Mat{M}$, which yields the decoded representation $\Mat{H}_{mc} \in \Space{R}^{\left| A_{mc} \right| \times d}$ as:
\begin{gather} \label{eq:decoder-mc}
    \Mat{H}_{mc} = \text{Transformer-Decoder}(\Mat{U}_{mc}, \Mat{M}).
\end{gather}
Notably, since the correctness of a answer candidate is invariant to its position, answer query is free of position encoding. 
Finally, we apply a linear projection on $\Mat{H}_{mc}$ to get the answer prediction $\hat{A}_{mc} \in \Space{R}^{\left| A_{mc} \right|}$,
\begin{gather}
    \hat{A}_{mc}=\text{Linear}(\Mat{H}_{mc}).
\end{gather}

\vspace{-15pt}
\subsubsection{Open-Ended QA}
Open-Ended setting provides $\left| A_{oe} \right|$ simple-form answer candidates (typically a single word) that are shared among all question instances, which makes the whole candidates set it too large to be processed as Multi-Choice setting. (\ie $\left| A_{oe} \right| \gg  \left| A_{mc} \right|$).
Instead, we take inspiration from DETR \cite{detr}, and initialize a single learnable embedding $\Mat{U}_{oe} \in \Space{R}^{d}$ as answer query.
Analogous to Multi-Choice setting, we feed $\Mat{U}_{oe}$ to the transformer decoder together with $\Mat{M}$ and acquire the decoded representation $\Mat{H}_{oe} \in \Space{R}^{d}$ similar to \cref{eq:decoder-mc}.
As a result, we obtain the prediction $\hat{A}_{oe} \in \Space{R}^{\left|A_{oe}\right|}$ by projecting $\Mat{H}_{oe}$ to the answer space $\Space{R}^{ \left|A_{oe}\right|}$ via a linear layer:
\begin{gather}
    \vspace{-2pt}
    \hat{A}_{oe}=\text{Linear}(\Mat{H}_{oe}).
        \vspace{-2pt}
\end{gather}

During training, we establish our objective on a cross-entropy loss.  For inference, the differentiable Top-K is replaced with vanilla hard Top-K for better efficiency. 

%% file: sec/4_exp.tex
\section{Experiments}\label{sec:exp}
\input{tab/dataset}

\noindent\textbf{Datasets:}
%
Experiments were conducted on four benchmarks to evaluate TranSTR's performance from different aspects. Specifically, the recent NExT-QA \cite{next-qa} and Causal-VidQA \cite{causalvid} datasets challenge models with complex VideoQA tasks using a Multi-Choice setting, which aims to test their temporal reasoning ability with complex causal and commonsense relations. In addition, MSVD-QA \cite{DBLP:conf/mm/XuZX0Z0Z17} and MSRVTT-QA \cite{DBLP:conf/mm/XuZX0Z0Z17} employ an Open-Ended setting and emphasize the description of video objects, activities, and their attributes. Their statistics are presented in Table \ref{tab:dataset}.

\vspace{5pt}
\noindent\textbf{Implementation Details:}
%
Following the convention established in \cite{VGT}, we sample each video as a sequence of $T$=16 frames, where each frame is encoded by a ViT-L \cite{vit} model that pre-trained on ImageNet-21k. To extract object-level features, we employ a Faster-RCNN \cite{faster-rcnn} model that pre-trained on the Visual Genome and detect $S$=20 objects in each frame.
For the textual encoding, we adopt a pretrained Deberta-base model \cite{deberta} to encode the question and answer. During training, we optimize the model using an Adam optimizer with a learning rate of 1e-5, and set the hidden dimension $d$ to 768. For the hyper-parameters, we set $K_f$=5 and $K_o$=12 for all datasets.

\vspace{5pt}
\noindent \textbf{Next}, we show our experimental results to answer the following questions:
\begin{itemize}[leftmargin=*]
\setlength\itemsep{-.20em}
    \item \textbf{Q1:} How is TranSTR compared with the SoTA?
    \item \textbf{Q2:} How effective are the proposed components?
    \item \textbf{Q3:} What learning pattern does the rationlizer capture?
\end{itemize}

\subsection{Main Result (Q1)}
In \cref{tab:main,tab:causal_vid}, we show that TranSTR outperforms SoTAs on all question types. Our observations are as follows:

\textbf{QA setting.}
Comparing the performance of TranSTR with state-of-the-art methods on four datasets, we observe that TranSTR achieves a greater improvement on Multi-Choice (NExT +5.8\% and Causal-Vid +6.8\%) compared to Open-End QA (MSVD +3.5\% and MSRVTT +3.4\%). This can be explained from two aspects:
(1) Unlike Open-Ended datasets that contain simple questions and short videos, Multi-Choice datasets (NExT and Causal-Vid) focus on complex VideoQA, in which composite question sentences with long videos and multiple objects (see \cref{tab:dataset}) makes the identifying and inspecting of critical scenes necessary. This aligns with TranSTR's design philosophy of removing redundancy and explicitly exposing critical elements. Therefore, TranSTR achieves a larger gain on complex VideoQA.
(2) In multi-choice QA, SoTA methods often append candidate answers to the question during encoding, which can create a spurious correlation between the negative answer and the question-irrelevant scene, leading to a false prediction. However, such an issue is less significant in open-ended QA, where each answer candidate is treated as a one-hot category without semantic meaning. Thus, the decoder of TranSTR brings extra benefits to the multi-choice setting, and result in larger gains compared to open-ended QA.

\textbf{Question-type.}
Based on the analysis of Multi-Choice datasets, we observe that the improvement in overall performance of TranSTR is largely due to the enhancement in answering composite questions (including Acc@C and Acc@T in NExT-QA, Acc@E and Acc@P and Acc@C in Causal-VidQA) that require deeper understanding such as causal relations and counterfactual thinking, compared to the descriptive question type (Acc@D:+1.8$\sim$2.7\%). This demonstrates TranSTR's outstanding reasoning ability for complex VideoQA tasks.
In particular, for Causal-VidQA, TranSTR shows a significant improvement in answering reason-based questions (Acc@P:AR +10.5\%, Acc@C:AR +8.3\%). Because questions of this type require the model to justify its prediction by selecting the correct evidence, which aligns with the concept of rationalization in TranSTR's design philosophy. Therefore, TranSTR's rationalization mechanism enables it to perform optimally in answering reason-based questions.
\input{tab/main}
\input{tab/causalvid}

\subsection{In-Depth Study (Q2)}
\subsubsection{Ablative Results} 
\input{tab/ablation}

We validate the key components of TranSTR by performing model ablation and discussing other implementation alternatives.
As shown in \cref{tab:ablation-loss}, we first study the effectiveness of TranSTR by removing both STR and decoder (``w/o STR \& decoder''), which induces a severe performance decline on every question type. 
Then, we conduct experiments to study STR. As a detailed breakdown test, we notice that reasoning with all frames without temporal rationalization (w/o ``TR'') will cause a performance drop.
A similar declination is also observed when spatial rationalization is erased (w/o ``SR''), that is, all objects on the selected frame are used for reasoning. 
Such performance drops are expected, because a large proportion of frames only contain a question-irrelevant scene, and the pretrained object detector will inevitably introduce noisy objects. These question-irrelevant contents, if not properly ruled out, will make the background overwhelm the causal information, due to its spurious correlation with the answer distractor. 
As a result, we witness a more significant performance drop when both temporal and spatial rationalization are removed (w/o ``STR'').
Next, we validate the effectiveness of our decoder design by adopting a conventional implementation that concatenates each answer candidate with the question before feeding it to the model. This variant, remarked as ``w/o decoder'', also caused a substantial performance drop, which highlights the importance of our video-text interaction pipeline in eliminating the background-distractor correlation.
Comparing the performance of (w/o ``STA'') and (w/o ``deocder'') to (``w/o STR \& decoder''), we show that removing both STR and decoder induce a more severe decline, which demonstrates that STR can coordinate well with the proposed decoder and their benefits are mutually reinforcing.
We also evaluate the importance of our MGR module by replacing it with average pooling. This variant, denoted as ``w/o MGR'', uses an average pooling to gather all objects on each frame and adds the pooled representation to the corresponding critical frames. We observed a significant performance drop when compared to the original TranSTR, which confirms the necessity of MGR in aggregating the multi-grain evidence for reasoning.
To validate that the STR indeed learns to focus on the critical elements instead of making random choices, we replace our differentiable top-K module, with a random K selection. As a result, the performance of ``Random K" drops drastically, which verifies the proposed STR is fully trainable to capture the answer information.
In addition, we also verify our choice of the differentiable module by replacing our perturbed maximum method with the ``SinkHorn Top-K'' \cite{sinkhorn}, and the results validate our implementation.

\subsubsection{Analysis on Complex VideoQA}
\input{tab/vt_obj}
\input{tab/decoder}

In \cref{tab:vt_obj }, we compare the results of TranSTR on the simple and complex VideoQA, where the test set of NExT-QA \cite{next-qa} is split by the length and object number of the source video, respectively. 
By calculating the accuracy within each group, we notice that all existing methods, as well as the TranSTR baseline (TranSTR without STR and proposed decoder), suffer from a performance drop when the video length exceeds 80 seconds (diff: -1.3\%$\sim$-2\%) or the video contains more than 5 objects (diff: -0.5\%$\sim$-1.9\%)). 
In contrast, TranSTR has alleviated this issue by explicitly ruling out redundant frames and noisy objects, thus resulting in even better performance on complex samples.

Similar to \cref{fig:1b}, we also investigate how the TranSTR performs on samples with different video lengths. In \cref{fig:vid_len}, we sort all test samples based on their video length and use a subset with the top percentage of longest videos to calculate accuracy. For example, 10\% on the x-axis denotes the accuracy of samples with the top 10\% longest videos, and 100\% denotes all samples considered.
Although the performance of TranSTR and baseline are initially comparable, the advantage of TranSTR becomes more pronounced as videos become longer. As a result, when the subset narrows down to the 10\% longest videos, we observe a difference in the accuracy of over 4\%.

\vspace{-5pt}
\subsubsection{Study of Decoder}
Our video-text interaction pipeline is able to cater to any SoTA methods without compromising their structure. Thus, we apply the proposed decoder to three VideoQA backbones by separating the answer candidates from the question and feeding them to our answer decoder. As shown in \cref{tab:decoder}, our decoder is able to consistently improve the performance of all backbones, validating the assumption that isolating the answer candidates from the video-text encoding can eliminate the spurious correlation between a negative answer and background scenes, resulting in favorable gains for the backbone models.
Moreover, our decoder design is also more efficient compared to the conventional implementation. In the traditional approach, a question needs to be fused with the video multiple times, with each time concatenating a different answer candidate. In contrast, our design only forwards the question once, as the answer candidates are introduced only after the video-question fusion, resulting in a much more efficient architecture.

\begin{figure}
	\begin{minipage}{0.48\linewidth}
		\centering
        \includegraphics[width=1\textwidth]{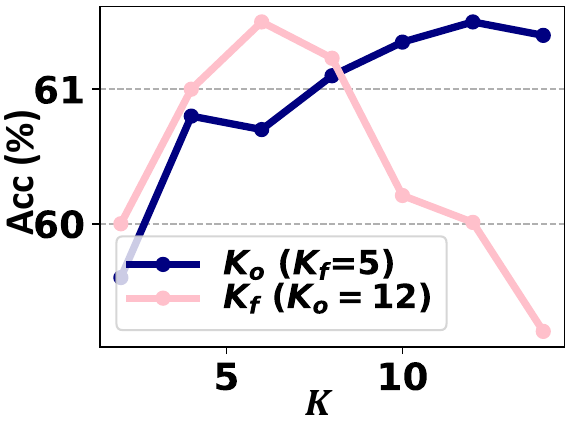}
         \vspace{-0.5cm}
        \captionof{figure}{Study of hyper-parameters.}
        \label{fig:k}
         \vspace{-0.5cm}
	\end{minipage}\hfill
	\begin{minipage}{0.48\linewidth}
		\centering
        \includegraphics[width=1\textwidth]{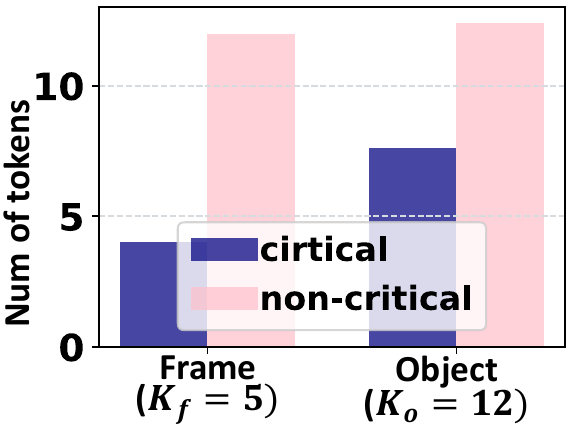}
         \vspace{-0.5cm}
        \captionof{figure}{Study of critical frames and objects.}            
	\label{fig:c}
         \vspace{-0.5cm}
	\end{minipage}
\end{figure}

\begin{figure*}[t]
  \centering
\scalebox{1.0}{
  \includegraphics[width=1.0\textwidth]{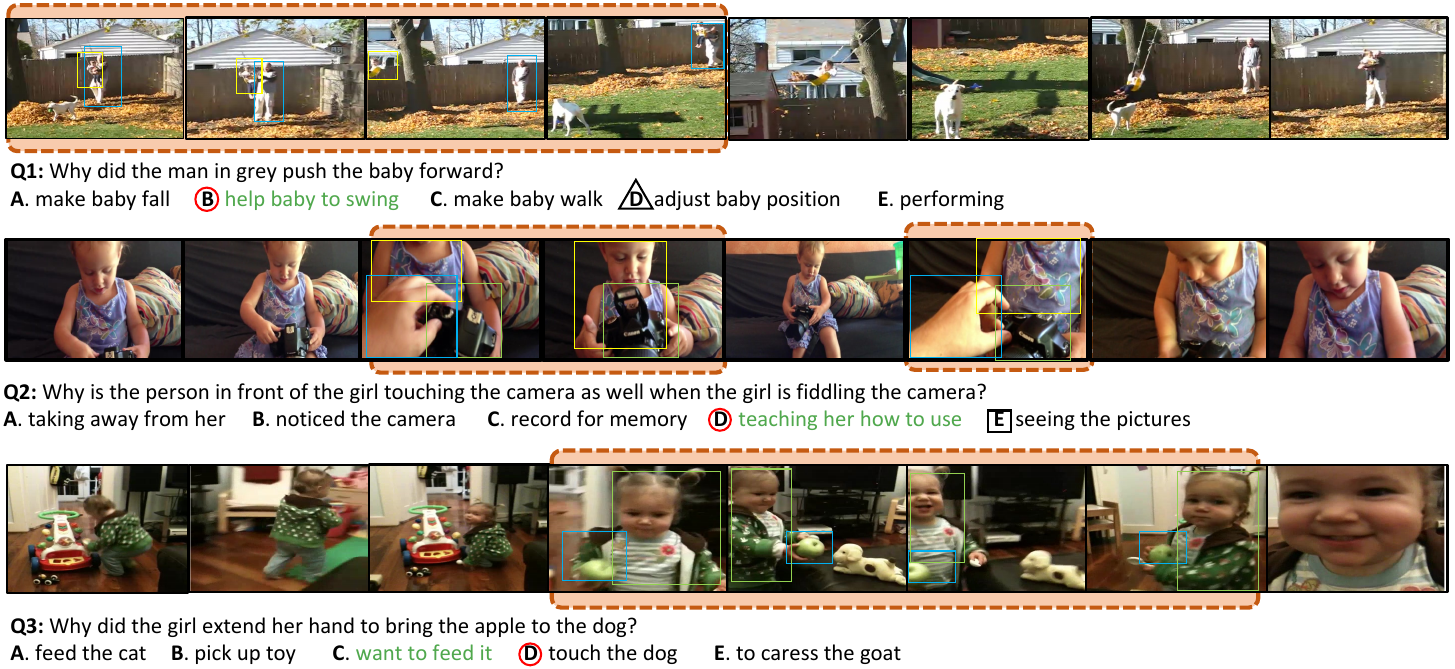}
  }
  	\vspace{-10pt}
  \caption{Case-Study on NExT-QA test set, the critical frames and objects are highlighted. Q1 and Q2 present the effect of the proposed STR and answer decoder, respectively. Q3 shows a failure case. The ground truth is colored in \green{green}. (
  \red{$\bigcirc$}:prediction of TranSTR,  $\bigtriangleup$: prediction of TranSTR w/o STR, $\Box$: prediction of TranSTR w/o decoder)}

  \label{fig:case_study}
 	\vspace{-13pt}
\end{figure*}

\vspace{-5pt}
\subsubsection{Study of Hyper-parameter}
To validate the sensitivity of TranSTR to the number of collected interactions, we conduct experiments with variations of $K_f$ and $K_o$ on NExT-QA. Without loss of generality, we tune $K_f$ ($K_o$) while setting $K_o$ = 12 ($K_f$ = 5). According to \cref{fig:k}, we observe that the performance of TranSTR varied mostly in the range of 61\% to 61.5\% under different combinations of hyperparameters, which demonstrated the effectiveness of TranSTR's adaptive design. However, we also notice a significant drop in some corner cases. When $K_f$ ($K_o$) is too small, the number of critical frames (objects) is limited, which hinders the performance as some visual evidence for answering is missing. Similarly, when $K_f$ is larger than 10, it introduces too much background, thus hurting the performance.

\subsection{Study of Critical Frames and Objects (Q3)}
\vspace{-3pt}
\noindent \textbf{Quantitative study.} 
To grasp the learning insight of TranSTR, we inspect the number of frames and objects that are collected as critical by the adaptive rationalization. 
Concretely, we draw the number of frames $C$ and non-critical frames $T$-$C$ in \cref{fig:k}. For the visual objects, since the number of critical objects $C_t$ varies according to frame content, we take the average of $C_t$ overall all critical frames, while leaving the rest objects as non-critical.
As a result, TranSTR can pinpoint a small group of tokens as critical tokens while leaving the rest as redundancy, which manifests the mass of question-irrelevant content in the original video, thus pointing the necessity of rationalization. 

\vspace{2pt}
\noindent \textbf{Qualitative study.}
To capture the learning pattern of TranSTR, we present some prediction results in \cref{fig:case_study} along with the identified frames and objects.
In general, TranSTR can locate very few indicative elements as visual evidence. 
In question 1, we show the effectiveness of the STR. In temporal selection, it rules out the environment scene and targets the first four "swing" frames as critical. Next, in the spatial selection, it excludes non-causal objects (\ie "dog") and focuses on the relation between question-relevant objects (\ie "man" and "baby"). By aggregating the critical elements in frames and objects, TranSTR successfully reaches the gold answer. As a comparison, when the STR is removed, the massive background overwhelms the salient reasoning pattern and leads to a false prediction.
Question 2 demonstrates the effect of our answer decoder. We can see that TranSTR targets three critical frames that encompass the question-referred ``person", while selecting ``camera" and ``girl" as critical objects to correctly infer the person's intention. However, when the decoder is removed, the prediction falls into negative answer ``E.seeing the pictures". This is because implementation without our answer decoder inevitably suffers from a spurious correlation between the negative answer ``E.seeing the pictures" and frames where the girl is actually seeing pictures, even though these frames are irrelevant to the question.
Lastly, we present a failure case in question 3, where TranSTR fails to capture the subtle difference between ``feed'' and ``touch'', although the critical visual elements are located, which leads to a false prediction of the girl's intention.

%% file: tab/dataset.tex
\setlength{\tabcolsep}{4pt}
\begin{table}[t!]
  \centering
  \small
  \caption{Dataset Statistics. MC and OE denote Multi-Choice and Open-Ended QA respectively.}
  \vspace{-0.1cm}
  \scalebox{0.8}{
    \begin{tabular}{lccccc}
    \toprule
    Dataset & Challenge & \#QA pair  & V-Len & Q-Len & QA \\
    \midrule
    NExT-QA & Causal \& Temporal & 48K   & 44s    &   11.6    & MC \\
    Causal-VidQA & Evidence \& Commonsense & 161K  & 9s     &   9.5    & MC \\
    MSVD-QA  & Description &   50K    & 10s    &    6.6   & OE \\
    MSRVTT-QA & Description & 244K  & 15s    &   7.4    & OE \\
    \bottomrule
    \end{tabular}
    }%
  \label{tab:dataset}%
 \vspace{-0.5cm}
\end{table}%

%% file: tab/main.tex
\setlength{\tabcolsep}{3.5pt}
\begin{table}[t!] 
  \centering
  \small
  \caption{Accuracy (\%) comparison on NExT-QA, MSVD-QA, and MSRVTT-QA. Acc@C, T, D, denote questions type of Causal, Temporal, and Descriptive in NExT-QA, respectively. The \textbf{best} and \underline{2nd best} results are highlighted.}
  \scalebox{0.8}{
    \begin{tabular}{l|c|ccc|c|c}
    \toprule
    \multirow{2}*{Methods} & 
    \multicolumn{4}{c|}{NExT-QA} & 
    \multirow{2}*{MSVD} & \multirow{2}*{MSRVTT} \\
    \cline{2-5} 
    ~  & Acc@All & Acc@C & Acc@T & Acc@D & ~ & ~ \\
    \midrule
    Co-Mem$\dagger$ \cite{gao2018motionappearance} & 48.5 & 45.9 & 50.0 & 54.4  & 34.6 & 35.3 \\
    HCRN \cite{hcrn}         & 48.9  & 47.1  & 49.3  & 54.0      & 36.1  & 35.6 \\
    HGA \cite{hga}         & 50.0 & 48.1  & 49.1  & 57.8      & 34.7  & 35.5 \\
    MSPAN \cite{mspan}           &  50.9  & 48.6    & 49.8     & 60.4       & 40.3  & 38.0 \\
    IGV \cite{IGV}   & 51.3 & 48.6  & 51.7  & 59.6   & 40.8 & 38.3 \\
    HQGA \cite{hqga}     & 51.8 & 49.0    & 52.3  & 59.4    & 41.2     & 38.6 \\
    EIGV \cite{EIGV}     & 52.9  & 51.2 & 51.5 & 61.0   & \underline{42.6}  & 39.3 \\
    VGT \cite{VGT}       & 53.7 & 51.6  & 51.9  & 63.7    & -     & \underline{39.7} \\

    VGT-PT \cite{VGT} &  \underline{55.7} & \underline{52.8}  & \underline{54.5}  & \underline{67.3}   & -     & - \\
    \midrule
    
    TranSTR  & \bf{61.5} & \bf{59.7}  & \bf{60.2}  & \bf{70.0}    & \bf{47.1}  & \bf{43.1} \\
    \textit{vs.} SoTA & +5.8 & +6.9 & +5.7 & +2.7  & +3.5 & +3.4 \\
    \bottomrule
    \end{tabular}}
  \label{tab:main}%
  \vspace{-0.5cm}
\end{table}%

%% file: tab/causalvid.tex
\setlength{\tabcolsep}{3pt}
\begin{table}[t!]
    \small
    \centering
    \caption{Accuracy (\%) comparison on Causal-VidQA. D: Description, E: Explanation, P: Prediction, C: Counterfactual. *: Reproduced result using official implementation.}
    \vspace{-0.5em}
    \scalebox{0.8}{
    \begin{tabular}{l|cccccccc|c}
    \toprule
    
    \multirow{2}*{Methods} & 
    \multirow{2}*{Acc@D} & 
    \multirow{2}*{Acc@E} & 
    \multicolumn{3}{c}{Acc@P} & \multicolumn{3}{c|}{Acc@C} & \multirow{2}*{Acc@All} \\
    \cline{4-9}
    ~ &~ & ~ & A & R & AR & A & R & AR &~ \\ 
    \midrule
    HCRN\cite{hcrn}  & 56.4 & 61.6 & 51.7 & 51.3 & \underline{32.6} & 51.6 & 53.4 & 32.7 & 48.1 \\
    HGA\cite{hga}  & 65.7 & 63.5 & 49.4 & 50.6 & 32.2 & 52.4 & 55.9 & 34.3 & 48.9 \\
    B2A\cite{park2021bridge}  & 66.2 & 62.9 & 49.0 & 50.2 & 31.2 & 53.3 & 56.3 & 35.2 & 49.1 \\
    VGT*\cite{VGT} & \underline{70.8} & \underline{70.3} & \underline{55.2} & \underline{56.9} & \underline{38.4} & \underline{61.0} & \underline{59.3} & \underline{42.0} & \underline{55.4}\\
    \midrule
    TranSTR & \bf{73.6} & \bf{75.8} & \bf{65.1} & \bf{65.0} & \bf{48.9} & \bf{68.6} & \bf{65.3} & \bf{50.3} & \bf{62.2}\\
     \textit{vs.} SoTA & +1.8 & +5.5 & +9.9 & +8.1 & +10.5 & +7.6 & +6.0 & +8.3 & +6.8 \\
    \bottomrule
    \end{tabular}
    }
    \label{tab:causal_vid}
    \vspace{-0.2cm}
\end{table}

%% file: tab/ablation.tex
\setlength{\tabcolsep}{9pt}
\begin{table}[t!]
  \centering
  \small
  \caption{Ablation Study}
  \vspace{-0.1cm}
    \scalebox{0.8}{
    \begin{tabular}{l|c|ccc}
    \toprule
    \multirow{2}*{Variants} & 
    \multicolumn{4}{c}{NExT-QA}\\
    \cline{2-5}
    ~  & Acc@All & Acc@C & Acc@T & Acc@D  \\
    \midrule
    TranSTR  & \bf{61.5} & \bf{59.7} & \bf{60.2} & \bf{70.0}  \\
    \midrule
    w/o STR \& decoder & 59.6 & 58.2  &  58.0  & 67.3  \\
    w/o STR & 60.3 & 59.1 & 58.3 & 67.9  \\
    $\;\;\;\;\;\;$ w/o TR & 60.8 & 59.4 &   58.6  &  69.5   \\
    $\;\;\;\;\;\;$ w/o SR & 60.7 & 59.6 & 58.2 & 69.0  \\
    w/o decoder & 60.1& 58.2   &   58.9   &  68.5  \\
    w/o MGR & 60.1 & 58.9 & 57.6 & 68.6  \\
    \midrule
    Random K & 54.6 & 53.6 &  51.6 & 64.0  \\
    SinkHorn Top-K & 61.0 & 59.4  &  59.7  &  68.7  \\


    \bottomrule
    \end{tabular}}%
  \label{tab:ablation-loss}%
  \vspace{-0.5cm}
\end{table}%

%% file: tab/vt_obj.tex
\setlength{\tabcolsep}{5pt}
\begin{table}[!t]
\small
\centering
\caption{Performance comparison, grouped by video length and object number. diff = Acc($\textgreater$ 80s) $-$ Acc( $\leq$80s)}.\label{tab:vt_obj }
  \scalebox{0.8}{
\begin{tabular}{l|ccc|ccc|c}
\toprule
\multirow{2}{*}{Model} &\multicolumn{3}{c|}{Video Length} &\multicolumn{3}{c|}{Object Number} &\multirow{2}{*}{Total} \\
\cline{2-7}
& $\leq$ 80s & $\textgreater$ 80s & diff($\uparrow$) & $\leq$ 5 & $\textgreater$ 5 & diff($\uparrow$) & \\
\midrule
EIGV \cite{EIGV} &53.3 &51.3 &-2 &53.3 &52.1 &-1.2 &52.9 \\
VGT \cite{VGT} &54.4 &52.2 &-2.2 &54.7 &52.8 &-1.9 &53.7 \\
VGT-PT \cite{VGT}  &55.8 &54.5 &-1.3 &56.4 &54.9 &-1.5 &55.7 \\
\midrule
Baseline &59.8 &58.7 &-1.1 &60.0 &59.2 &-0.8 & 59.6 \\
TranSTR &61.4 &62.4 & \bf{+1} &61.2 &61.9 & \bf{+0.7} &61.5 \\
\bottomrule
\end{tabular}}
\end{table}

%% file: tab/decoder.tex
          



\begin{figure}
	\begin{minipage}{0.52\linewidth}
		\centering
        \vspace{-7pt}
        \includegraphics[width=0.97\textwidth]{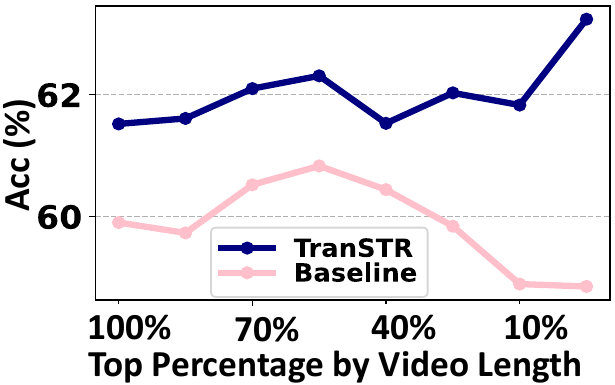} 
        \vspace{-5pt}
        \captionof{figure}{Acc by video length}
        \label{fig:vid_len}
	\end{minipage}\hfill
	\begin{minipage}{0.45\linewidth}

		\centering
            \small
		\resizebox{0.95\textwidth}{!}{%
            
		    \begin{tabular}{l|cc}
        \toprule
        \multirow{2}*{Methods} & 
        \multicolumn{2}{c}{Decoder} \\
        ~ & w/o & w \\
        \midrule
        MSPAN \cite{mspan} &50.9 &52.7 \\
        EIGV \cite{EIGV} &51.3 &53.3 \\
        TranSTR &60.1 &61.5 \\
        \bottomrule
    \end{tabular}
                }
                \captionof{table}{Apply our decoder to SoTAs.}
                \label{tab:decoder}
	\end{minipage}
 \vspace{-0.5cm}
\end{figure}




%% file: sec/5_conclusion.tex
\section{Conclusion} \label{sec:conclusion}
\vspace{-3pt}
For the first time, this paper addresses complex VideoQA, where long multi-object video and hard answer distractors have crippled existing methods, owing to their incapacity in handling massive visual backgrounds and modeling hard distractor answers. We then propose STR to adaptively trim off question-irrelevant scenes, and further develop a novel answer decoding scheme that coordinates with STR to overcome the spurious correlation resulted from distractor-background interaction.
Instantiating this pipeline with transformer architecture, we show our method, TranSTR, achieves significant improvements over SoTAs, especially on complex VideoQA tasks. We hope our success can shed light on answering questions in the context of long videos with multiple objects.


%% file: main.bbl
\begin{thebibliography}{10}\itemsep=-1pt

\bibitem{pertub}
Quentin Berthet, Mathieu Blondel, Olivier Teboul, Marco Cuturi, Jean-Philippe
  Vert, and Francis Bach.
\newblock Learning with differentiable pertubed optimizers.
\newblock {\em NeurIPS}, pages 9508--9519, 2020.

\bibitem{detr}
Nicolas Carion, Francisco Massa, Gabriel Synnaeve, Nicolas Usunier, Alexander
  Kirillov, and Sergey Zagoruyko.
\newblock End-to-end object detection with transformers.
\newblock In Andrea Vedaldi, Horst Bischof, Thomas Brox, and Jan{-}Michael
  Frahm, editors, {\em ECCV}, pages 213--229, 2020.

\bibitem{rationalization-robustness}
Howard Chen, Jacqueline He, Karthik Narasimhan, and Danqi Chen.
\newblock Can rationalization improve robustness?
\newblock In {\em NAACL}, pages 3792--3805, 2022.

\bibitem{hostr}
Long~Hoang Dang, Thao~Minh Le, Vuong Le, and Truyen Tran.
\newblock Hierarchical object-oriented spatio-temporal reasoning for video
  question answering.
\newblock In {\em IJCAI}, pages 636--642, 2021.

\bibitem{devlin2018bert}
Jacob Devlin, Ming-Wei Chang, Kenton Lee, and Kristina Toutanova.
\newblock Bert: Pre-training of deep bidirectional transformers for language
  understanding.
\newblock {\em arXiv preprint arXiv:1810.04805}, 2018.

\bibitem{vit}
Alexey Dosovitskiy, Lucas Beyer, Alexander Kolesnikov, Dirk Weissenborn,
  Xiaohua Zhai, Thomas Unterthiner, Mostafa Dehghani, Matthias Minderer, Georg
  Heigold, Sylvain Gelly, Jakob Uszkoreit, and Neil Houlsby.
\newblock An image is worth 16x16 words: Transformers for image recognition at
  scale.
\newblock In {\em ICLR}, 2021.

\bibitem{DBLP:conf/cvpr/DuaKB21}
Radhika Dua, Sai~Srinivas Kancheti, and Vineeth~N. Balasubramanian.
\newblock Beyond {VQA:} generating multi-word answers and rationales to visual
  questions.
\newblock In {\em CVPR}, pages 1623--1632, 2021.

\bibitem{fan2019heterogeneous}
Chenyou Fan, Xiaofan Zhang, Shu Zhang, Wensheng Wang, Chi Zhang, and Heng
  Huang.
\newblock Heterogeneous memory enhanced multimodal attention model for video
  question answering.
\newblock In {\em CVPR}, pages 1999--2007, 2019.

\bibitem{fu2021violet}
Tsu-Jui Fu, Linjie Li, Zhe Gan, Kevin Lin, William~Yang Wang, Lijuan Wang, and
  Zicheng Liu.
\newblock Violet: End-to-end video-language transformers with masked
  visual-token modeling.
\newblock {\em arXiv preprint arXiv:2111.12681}, 2021.

\bibitem{gao2018motionappearance}
Jiyang Gao, Runzhou Ge, Kan Chen, and Ram Nevatia.
\newblock Motion-appearance co-memory networks for video question answering.
\newblock In {\em CVPR}, pages 6576--6585, 2018.

\bibitem{mspan}
Zhicheng Guo, Jiaxuan Zhao, Licheng Jiao, Xu Liu, and Lingling Li.
\newblock Multi-scale progressive attention network for video question
  answering.
\newblock In {\em ACL}, pages 973--978, 2021.

\bibitem{deberta}
Pengcheng He, Xiaodong Liu, Jianfeng Gao, and Weizhu Chen.
\newblock Deberta: decoding-enhanced bert with disentangled attention.
\newblock In {\em ICLR}, 2021.

\bibitem{jang2017tgif}
Yunseok Jang, Yale Song, Youngjae Yu, Youngjin Kim, and Gunhee Kim.
\newblock Tgif-qa: Toward spatio-temporal reasoning in visual question
  answering.
\newblock In {\em CVPR}, pages 2758--2766, 2017.

\bibitem{hga}
Pin Jiang and Yahong Han.
\newblock Reasoning with heterogeneous graph alignment for video question
  answering.
\newblock In {\em AAAI}, pages 11109--11116, 2020.

\bibitem{DBLP:conf/icml/KumarIOIBGZPS16}
Ankit Kumar, Ozan Irsoy, Peter Ondruska, Mohit Iyyer, James Bradbury, Ishaan
  Gulrajani, Victor Zhong, Romain Paulus, and Richard Socher.
\newblock Ask me anything: Dynamic memory networks for natural language
  processing.
\newblock In {\em ICML}, volume~48, pages 1378--1387, 2016.

\bibitem{hcrn}
Thao~Minh Le, Vuong Le, Svetha Venkatesh, and Truyen Tran.
\newblock Hierarchical conditional relation networks for video question
  answering.
\newblock In {\em CVPR}, pages 9969--9978, 2020.

\bibitem{lei2021less}
Jie Lei, Linjie Li, Luowei Zhou, Zhe Gan, Tamara~L Berg, Mohit Bansal, and
  Jingjing Liu.
\newblock Less is more: Clipbert for video-and-language learning via sparse
  sampling.
\newblock In {\em CVPR}, pages 7331--7341, 2021.

\bibitem{causalvid}
Jiangtong Li, Li Niu, and Liqing Zhang.
\newblock From representation to reasoning: Towards both evidence and
  commonsense reasoning for video question-answering.
\newblock In {\em CVPR}, pages 21241--21250, 2022.

\bibitem{EIGV}
Yicong Li, Xiang Wang, Junbin Xiao, and Tat{-}Seng Chua.
\newblock Equivariant and invariant grounding for video question answering.
\newblock {\em CoRR}, abs/2207.12783, 2022.

\bibitem{IGV}
Yicong Li, Xiang Wang, Junbin Xiao, Wei Ji, and Tat-Seng Chua.
\newblock Invariant grounding for video question answering.
\newblock In {\em {CVPR}}, pages 2928--2937, 2022.

\bibitem{sinkhorn}
Gonzalo~E. Mena, David Belanger, Scott~W. Linderman, and Jasper Snoek.
\newblock Learning latent permutations with gumbel-sinkhorn networks.
\newblock In {\em ICLR}, 2018.

\bibitem{DBLP:conf/cvpr/ParkHARSDR18}
Dong~Huk Park, Lisa~Anne Hendricks, Zeynep Akata, Anna Rohrbach, Bernt Schiele,
  Trevor Darrell, and Marcus Rohrbach.
\newblock Multimodal explanations: Justifying decisions and pointing to the
  evidence.
\newblock In {\em CVPR}, pages 8779--8788, 2018.

\bibitem{park2021bridge}
Jungin Park, Jiyoung Lee, and Kwanghoon Sohn.
\newblock Bridge to answer: Structure-aware graph interaction network for video
  question answering, 2021.

\bibitem{pgat}
Liang Peng, Shuangji Yang, Yi Bin, and Guoqing Wang.
\newblock Progressive graph attention network for video question answering.
\newblock In {\em {ACM} {MM}}, 2021.

\bibitem{faster-rcnn}
Shaoqing Ren, Kaiming He, Ross~B. Girshick, and Jian Sun.
\newblock Faster {R-CNN:} towards real-time object detection with region
  proposal networks.
\newblock In {\em NeurIPS}, pages 91--99, 2015.

\bibitem{DBLP:conf/kdd/Ribeiro0G16}
Marco~T{\'{u}}lio Ribeiro, Sameer Singh, and Carlos Guestrin.
\newblock "why should {I} trust you?": Explaining the predictions of any
  classifier.
\newblock In Balaji Krishnapuram, Mohak Shah, Alexander~J. Smola, Charu~C.
  Aggarwal, Dou Shen, and Rajeev Rastogi, editors, {\em KDD}, pages 1135--1144,
  2016.

\bibitem{vidvrd}
Xindi Shang, Tongwei Ren, Jingfan Guo, Hanwang Zhang, and Tat{-}Seng Chua.
\newblock Video visual relation detection.
\newblock In {\em ACM MM}, pages 1300--1308, 2017.

\bibitem{DIR}
Yingxin Wu, Xiang Wang, An Zhang, Xiangnan He, and Tat{-}Seng Chua.
\newblock Discovering invariant rationales for graph neural networks.
\newblock In {\em ICLR}, 2022.

\bibitem{next-qa}
Junbin Xiao, Xindi Shang, Angela Yao, and Tat{-}Seng Chua.
\newblock Next-qa: Next phase of question-answering to explaining temporal
  actions.
\newblock In {\em CVPR}, pages 9777--9786, 2021.

\bibitem{hqga}
Junbin Xiao, Angela Yao, Zhiyuan Liu, Yicong Li, Wei Ji, and Tat-Seng Chua.
\newblock Video as conditional graph hierarchy for multi-granular question
  answering.
\newblock In {\em AAAI}, pages 2804--2812, 2022.

\bibitem{VGT}
Junbin Xiao, Pan Zhou, Tat-Seng Chua, and Shuicheng Yan.
\newblock Video graph transformer for video question answering.
\newblock In {\em ECCV}, pages 39--58. Springer, 2022.

\bibitem{DBLP:conf/mm/XuZX0Z0Z17}
Dejing Xu, Zhou Zhao, Jun Xiao, Fei Wu, Hanwang Zhang, Xiangnan He, and Yueting
  Zhuang.
\newblock Video question answering via gradually refined attention over
  appearance and motion.
\newblock In {\em ACM MM}, pages 1645--1653, 2017.

\bibitem{yang2021just}
Antoine Yang, Antoine Miech, Josef Sivic, Ivan Laptev, and Cordelia Schmid.
\newblock Just ask: Learning to answer questions from millions of narrated
  videos.
\newblock In {\em Proceedings of the IEEE/CVF International Conference on
  Computer Vision}, pages 1686--1697, 2021.

\bibitem{zellers2021merlot}
Rowan Zellers, Ximing Lu, Jack Hessel, Youngjae Yu, Jae~Sung Park, Jize Cao,
  Ali Farhadi, and Yejin Choi.
\newblock Merlot: Multimodal neural script knowledge models.
\newblock {\em Advances in Neural Information Processing Systems},
  34:23634--23651, 2021.

\bibitem{DBLP:conf/cvpr/ZhangYMW19}
Quanshi Zhang, Yu Yang, Haotian Ma, and Ying~Nian Wu.
\newblock Interpreting cnns via decision trees.
\newblock In {\em CVPR}, pages 6261--6270, 2019.

\bibitem{zhong2022video}
Yaoyao Zhong, Junbin Xiao, Wei Ji, Yicong Li, Weihong Deng, and Tat-Seng Chua.
\newblock Video question answering: Datasets, algorithms and challenges.
\newblock In {\em Proceedings of the 2022 Conference on Empirical Methods in
  Natural Language Processing}, pages 6439--6455, 2022.

\end{thebibliography}
